\documentclass[journal]{IEEEtran}
\usepackage{amsmath,amsfonts}
\usepackage{algorithmic}
\usepackage{array}
\usepackage[caption=false,font=normalsize,labelfont=sf,textfont=sf]{subfig}
\usepackage{textcomp}
\usepackage{stfloats}
\usepackage{url}
\usepackage{verbatim}
\usepackage{graphicx}
\usepackage{booktabs}
\usepackage{multirow}
\usepackage{multicol}
\usepackage{array}
\usepackage{caption}
\usepackage{makecell}
\usepackage[ruled]{algorithm2e}
\usepackage{xcolor,colortbl}
\usepackage[shortlabels]{enumitem}
\usepackage{diagbox}
\def\BibTeX{{\rm B\kern-.05em{\sc i\kern-.025em b}\kern-.08em
    T\kern-.1667em\lower.7ex\hbox{E}\kern-.125emX}}
\usepackage{balance}

\begin{document}

\title{
Wide Flat Minimum Watermarking for Robust Ownership Verification of GANs}
\author{Jianwei Fei, Zhihua Xia, \IEEEmembership{Member, IEEE}, Benedetta Tondi, \IEEEmembership{Member, IEEE}, and Mauro Barni, \IEEEmembership{Fellow, IEEE}
\thanks{

Jianwei Fei is with the College of Cyber Security, Jinan University, Guangzhou, China, and the University of Siena, Siena, Italy (e-mail: fjw826244895@163.com).

Zhihua Xia is with the College of Cyber Security, Jinan University, Guangzhou, China, and also with the Engineering Research Center of Digital Forensics, Nanjing University of Information Science and Technology, Nanjing, China (e-mail: xia\_zhihua@163.com).

Benedetta Tondi and Mauro Barni are with the University of Siena, Siena, Italy (e-mail: benedettatondi@gmail.com, barni@dii.unisi.it).

}}

\markboth{Journal of \LaTeX\ Class Files,~Vol.~18, No.~9, September~2020}%
{How to Use the IEEEtran \LaTeX \ Templates}

\maketitle

\begin{abstract}
We propose a novel multi-bit box-free watermarking method for the protection of Intellectual Property Rights (IPR) of GANs with improved robustness against white-box attacks like fine-tuning, pruning, quantization, and surrogate model attacks. The watermark is embedded by adding an extra watermarking loss term during GAN training, ensuring that the images generated by the GAN contain an invisible watermark that can be retrieved by a pre-trained watermark decoder. In order to improve the robustness against white-box model-level attacks, we make sure that the model converges to a wide flat minimum of the watermarking loss term, in such a way that any modification of the model parameters does not erase the watermark. To do so, we add  random noise vectors to the parameters of the generator and require that the watermarking loss term is as invariant as possible with respect to the presence of noise. This procedure forces the generator to converge to a wide flat minimum of the watermarking loss. The proposed method is architecture- and dataset-agnostic, thus being applicable to many different generation tasks and models, as well as to CNN-based image processing architectures. We present the results of extensive experiments showing that the presence of the watermark has a negligible impact on the quality of the generated images, and proving the superior robustness of the watermark against model modification and surrogate model attacks.
\end{abstract}


\section{Introduction}
\IEEEPARstart{I}{n} the past decade, deep learning has rapidly gained popularity due to its superior performance compared to previous approaches and even humans, in various tasks and application scenarios, including computer vision, healthcare, finance, and manufacturing. The success of deep learning relies on well-designed and trained Deep Neural Networks (DNN), whose training requires the availability of massive amounts of labeled data, computational resources, and expertise. One concern is that attackers who are unable to train such models themselves may steal or use them without respecting the license terms under which they are made available. Therefore, it is important that means to protect the IPR of DNN models are developed.

Inspired by the use of multimedia watermarking for IPR protection~\cite{BB04}, several researchers have proposed DNN watermarking as a solution to protect the IPR of DNN models~\cite{Li2021b}. With DNN watermarking the ownership information is embedded directly or indirectly into the model parameters. Subsequently, the ownership can be verified by matching the extracted watermark with the owner's watermark.

Generally speaking, DNN watermarking methods can be split into 3 categories according to the way the watermark is extracted from the network~\cite{Li2021b}. In white-box methods, the watermark is read from the internal parameters/status of the network, thus the decoder is assumed to have full access to the model. In black-box methods, the decoder can only access the final output of the network and the watermark is recovered by querying the model with some specific inputs and looking at the output corresponding to those inputs. Eventually, for generative or image processing networks, the watermark can be read from any arbitrary sample generated by the watermarked model (box-free methods). Box-free methods do not require any knowledge about the model parameters, or the use of key inputs to interact with the model. For this reason, whenever applicable, e.g., with GANs or any other kinds of generative models, box-free methods provide a flexible way to prove the ownership of a network by looking only at the output samples it produces.

Early research has focused mainly on black-box and white-box methods, however protecting the IPR of GANs (and other image processing and generative models) by relying on box-free watermarking is a promising approach that deserves to be investigated further. So far, there have been only a few works on GAN watermarking, including black-box methods~\cite{Ong2021,Fei2022,qiao2023novel}, along with some box-free methods~\cite{Yu2021,yuresponsible,ruan2023intellectual}. Most existing black-box GAN watermarking methods are zero-bit schemes whereby it is only possible to determine whether the analyzed model contains a given known watermark or not. On the contrary, most box-free methods proposed so far are multi-bit watermarking schemes, where the watermark bits can be extracted from any output of the watermarked model without knowing them in advance. In this case, the authentication process involves matching the extracted watermark to the bits identifying the hypothesized model owner.
Due to their flexibility and large payload, multi-bit box-free watermarking schemes are in general preferable to black-box and white-box methods for GANs. However, this is not the case when we consider robustness against watermark removal attacks, especially model-level removal attacks. While existing black-box and white-box methods have shown good robustness against model-level watermark removal attacks, box-free methods are not robust enough to be used in practical scenarios.

In this paper, we propose a robust, box-free, multi-bit GAN watermarking method based on the concept of Wide Flat Minimum (WFM), which significantly improves the robustness against model-level watermark removal attacks, including fine-tuning, pruning, quantization, and surrogate model attacks, with respect to prior methods belonging to the same category.

In the embedding process, a pre-trained image watermarking decoder is applied to the generated images, and the generator is trained in such a way that the decoder correctly reads a pre-defined watermark from the generated images. To do so, we add a dedicated watermarking loss term to the loss function used to train the GAN. In order to overcome the robustness limitations of current GAN watermarking schemes, we force the model to converge to a WFM of the watermarking loss. In this way, the watermarking loss tends to be invariant to small and even moderate modifications of the model parameters, hence ensuring that the watermark can survive model manipulations, like pruning, quantization, and, most importantly, fine-tuning. To the best of our knowledge, our scheme is the first multi-bit GAN watermarking scheme achieving strong robustness against white-box model-level attacks, including fine-tuning and surrogate model attacks, without impairing the quality of the generated images. Such a result is proven by means of extensive experiments carried out on several GAN and CNN architectures addressing different tasks. Note that although in this paper we mainly focus on GAN architectures, our method is applicable to a wide range of generative models that produce images as output.

With the above ideas in mind, the contributions of this paper can be summarized as follows:
\begin{itemize}
	\item We present a new, multi-bit, box-free GAN watermarking method with strong robustness against model-level attacks, including fine-tuning and surrogate model attacks. To the best of our knowledge, this is the first multi-bit box-free GAN watermarking achieving such a level of robustness.
	\item Our method is architecture- and task-agnostic and thus can be used on a variety of image generation or processing tasks, including image translation, image synthesis, super-resolution, etc.
	\item We prove, under a rigorous hypothesis testing framework, that the proposed method can satisfy the watermark accuracy requirements for model ownership authentication under the considered attacks.
\end{itemize}

The rest of this paper is organized as follows. A review of related works is carried out in Section~\ref{sec: RELATED WORK}. The problem definition and threat model are presented in Section~\ref{sec: Problem definition and Threat Model}. Then, the proposed method is described in Section~\ref{sec: THE PROPOSED METHOD}. The experimental methodology and the results of the experiments are presented in Sections~\ref{sec:SETTINGS} and~\ref{sec: EXPERIMENTS} respectively. The paper ends in Section~\ref{sec: CONCLUSION} with a summary of our findings and some concluding remarks.

\section{RELATED WORK}
\label{sec: RELATED WORK}
According to a widely accepted taxonomy~\cite{Li2021b}, DNN watermarking methods can be categorized into white-box, black-box, and box-free methods. Moreover, depending on the information carried out by the watermark and the way the watermark information is extracted from the hosting network, DNN watermarking methods can be classified as multi-bit and zero-bit methods. In multi-bit watermarking, the watermark is a sequence of bits that can be extracted from the watermarked model without knowing it in advance. On the other hand, with zero-bit watermarking the watermark detector can only determine whether a known watermark is present in the model or not. In this section, we first give a general review of existing watermarking methods, then we describe existing box-free GAN watermarking methods in more detail.

Uchida~\textit{et al.}~\cite{Uchida2017} were the first to use digital watermarking for IPR protection of CNNs adopting a white-box setting. In their approach, a regularization term is incorporated in the loss used to train the target model, in order to impose a statistical bias on the parameters. This statistical bias ensures that a known binary watermark can be extracted from the model parameters using a preset projection matrix. Subsequently, numerous works on white-box watermarking have been proposed~\cite{Li2021a,LeMerrer2020,DarvishRouhani2019,Li2021,fan2019rethinking,fan2021deepipr}. White-box watermarking directly embeds watermark bits into the statistics of the model parameters, thus requiring white-box access for watermark detection/decoding. This limitation hinders the practical use of white-box watermarking, leading to a shift of focus toward black-box watermarking.

Black-box DNN watermarking methods assume that the parameters of the target model cannot be accessed, but the verifier is capable of querying the model with specific inputs, referred to as key inputs or trigger inputs. The presence of the watermark can be detected by analyzing the output of the watermarked model in correspondence to the key inputs. Existing research focuses on the construction of the key inputs and on the strategies to train the network in such a way to enforce specific outputs in correspondence to the key inputs~\cite{Zhang2018,li2020protecting, Zhong2020a, LeMerrer2020,chen2019blackmarks}. Black-box watermarking methods typically adopt the zero-bit watermarking paradigm, which lacks the flexibility necessary for applications like fingerprinting and traitor tracing. In addition, even if with black-box methods watermark extraction does not require white-box access to the inspected model, the security of the watermark may be at risk due to the necessity of exposing the key inputs during the watermark verification procedure~\cite{Li2019, li2020protecting}.

\subsection{Box-free Watermarking}
\label{sec: Box-free DNN Watermarking}
Unlike classification models, generative models, particularly those based on GANs, generate images (or other kinds of documents) with a large entropic content, which can naturally serve as carriers for multi-bit watermarks. Therefore, researchers have proposed box-free watermarking methods specifically designed for generative models, utilizing the images generated by the network as watermark carriers.

Yu~\textit{et al.}~\cite{Yu2021} were the first to propose a multi-bit box-free GAN watermarking method. They first embed a certain watermark into the training data by using a DNN image watermarking network (specifically Stegastamp~\cite{Tancik2020}), then use the watermarked data to train the target GAN. They demonstrate that the same watermark contained in the training data can be extracted from any image generated by the GAN. Yu~\textit{et al.}~\cite{yuresponsible} also proposed a scalable fingerprinting method inspired by the style-based encoder used in styleGAN~\cite{Karras2019}. By parameterizing the convolution kernels of the generator with different watermarks, and by forcing the generator to produce images with the input watermark, their method allows efficient and scalable generation of numerous models with distinct watermarks, yet maintaining identical functionalities.
%
%
Wu~\textit{et al.}~\cite{Wu2020} utilize images as watermarks and optimize the generator and a decoder with a combined loss function, ensuring that the decoder outputs only specific watermark images when receiving images generated by the target generator. Similarly, Fei~\textit{et al.}~\cite{Fei2022} have proposed a method to watermark GANs by enforcing the presence of a specific watermark within the generated images. Their method utilizes a pre-trained image watermarking decoder to impose an additional watermark loss term during training.
Considering that with box-free methods, the watermark is extracted from the generated images, it is crucial to take into account the robustness of the watermark against image processing attacks. In most cases, such a robustness is achieved by introducing a noise layer to simulate image processing attacks during watermark embedding.

In addition to image processing attacks, it is also important to consider the robustness against model-level attacks, as usually done for both white-box and black-box methods. This aspect, however, is rarely investigated in papers proposing box-free methods. Typically, if an attacker obtains white-box access to the target model, e.g. by stealing it or by violating the license agreement, he/she can carry out model-level watermark removal attacks, such as fine-tuning, pruning, and quantization, before utilizing the model. The goal of these attacks is to modify the watermarked GAN to let it generate images that do not contain the expected watermark, thereby disabling the ownership verification functionality. Even if the attackers have only black-box access to the watermarked model, he/she can still carry out a surrogate model attacks, by training a similar model using the input-output pairs of the target model.
These watermark removal attacks represent a serious threat to box-free GAN watermarking methods, however they have rarely been considered in existing research. In this work, we propose a novel GAN watermarking scheme for IPR protection achieving good robustness against both image-level and model-level attacks, significantly outperforming state-of-the-art methods in the latter case.

\section{Problem definition and Threat Model}
\label{sec: Problem definition and Threat Model}

In this section, we introduce the notations used throughout the paper, describe the threat model adopted in our work, and detail the hypothesis testing problem underlying the watermark-based owner verification protocol.

\subsection{Notation}
A GAN consists of a generator $G_\theta$ parameterized by the set of weights and biases $\theta$ and a discriminator $D_{gan}$. As an example, we may consider an image-transfer GAN taking as input image $x$ belonging to the input domain and mapping it into an image $G_\theta(x)$ belonging to the output domain. Our objective is to watermark $G_\theta$ in such a way that any image $G_\theta(x)$ generated by the GAN carries an invisible bit string (the watermark) $w \in \{0,1\}^n$ of length $n$, The string $w$ can be extracted by a watermark decoder $D_w$.
More formally, according to the box-free watermark retrieval paradigm, given any image $y = G_\theta(x)$ produced by the GAN, we let
\begin{equation}
    \hat{w} = D_w(y).
\label{eq.wat_extr}
\end{equation}
Then, the extracted watermark $\hat{w}$ is compared with the watermark bit sequence we are looking for (that is $w$). If $D_w(G_\theta(x))$ is equal to $w$ or the difference between $w$ and $\hat{w}$ is smaller than a threshold, we say that $G_\theta$ contains $w$ and the ownership of $G_\theta$ is proven. Given that both the model and the generated images may undergo a number of intentional or non-intentional modifications, we require that the watermark be retrievable from the modified images and even by the images produced by a modified version of the generator (say $G_{\theta'}$).
\begin{figure}[h]
	\centering
	\includegraphics[width=0.95\linewidth]{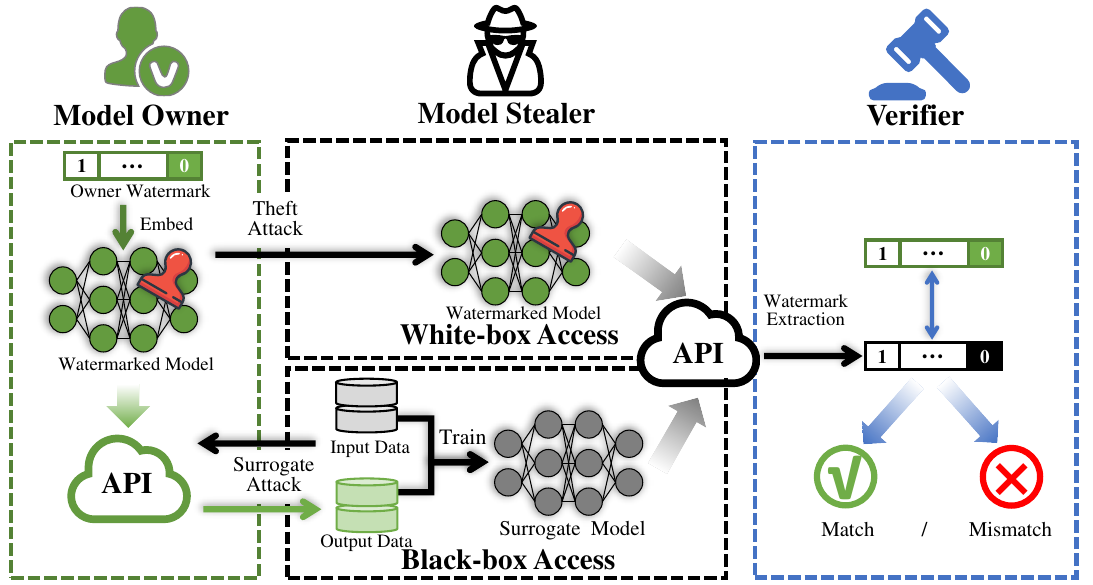}
	\caption{The threat model considered in our work.}
	\label{fig:Threat Model}
\end{figure}

\subsection{Threat Model}
\label{subsection: Threat Model}

The threat model considered in this paper (see Fig. \ref{fig:Threat Model}) involves three entities: the model owner, an attacker, or model stealer, and the verifier, who can be a third party or the model owner himself. The owner makes the model available through an API allowing the users to query the model and download the generated images. To avoid that the model is stolen and re-used without authorization, the owner embeds a watermark within it, to make it possible to prove the ownership of the model itself or that of any model derived from it, e.g. after fine-tuning. The ownership of the model is proven by verifying that the images it produces contain the watermark of the owner. The presence of the watermark within an image can also be used to prove that the image has been generated by the stolen model or by a model derived from it.

The goal of the attacker (stealer) is to steal the target model and use it for its own purposes without being traced. Specifically, we establish 2 scenarios based on the accessibility of the target model.
\begin{enumerate}[a), leftmargin=*]
	\item In the first scenario, the stealer may be an internal attacker or a skilled hacker who manages to create his own copy of the model. The stealer may also attempt to remove the watermark by modifying the model and/or by post-processing the generated images.
	\item In the second scenario, the attacker can access the model in a black-box modality, by querying it through the API made available by the owner. In this case, the attacker may exploit the input-output pairs obtained through the API to train a new {\em surrogate} model and use it for his own goals.
\end{enumerate}

We assume that after the attack, the verifier and the model owner can not access the suspicious model directly, rather they have access to the images produced by the model either through an API or by gathering such images on the Internet, through social networks, or by any other means. The objective of the verifier is to extract the watermark from the images produced by the suspicious model to verify whether it corresponds to the model of the legitimate owner or is derived from it.

\subsection{Ownership Verification}
\label{subsection: Ownership Verification}

Ownership verification is accomplished by running a hypothesis test on the presence of the watermark within the images generated by the suspicious model. To begin with, we assume that the verifier has access to only one image produced by the model, namely $y = G_\theta(x)$. Note that, following the box-free paradigm, we assume that the verifier does not know, or cannot choose the input $x$ used to generate $y$. The null hypothesis $H_0$ states that $y$ is not generated by the inspected model, while the alternative hypothesis $H_1$ states that $y$ is generated by the model.

Under $H_0$, the bits $\hat{w}$ obtained by applying the watermark decoder are independent of the owner's watermark sequence $w$, so the number of matching bits $k$ between $w$ and $\hat{w}$ follows a binomial distribution with matching probability $p=0.5$:
\begin{equation}
	P(k) = \binom{n}{k} 0.5^n,
	\label{eq:binomialH0}
\end{equation}
where $n$ is the length of the watermark. By assuming that the presence of the watermark is verified when at least $T$ bits of $\hat{w}$ match $w$, the false detection probability ($P_f$) is equal to :
\begin{equation}
	P_f = \sum_{k=T}^n \binom{n}{k} 0.5^n.
	\label{eq:nonwatermarked model}
\end{equation}

Given $n$ and the maximum allowed false detection probability, the above equation permits to compute the detection threshold. The threshold values corresponding to $n=100$ and various $P_f$ are reported in Table~\ref{tab:P_f and k}.

\begin{table}[htbp]
	\caption{Detection thresholds $T$ for different values of $P_f$ ($n= 100$).}
	\label{tab:P_f and k}
    \centering
    \begin{tabular}{c|cccccc}
    \toprule
    $P_f$   & 0.05     & 0.01  & $10^{-3}$   & $10^{-4}$ & $10^{-5}$ & $10^{-6}$ \\ \midrule
    $T$  & 55   & 60   & 65 & 68 & 71 & 74\\
    \bottomrule
    \end{tabular}
\end{table}

To calculate the missed detection probability ($P_m$) under $H_1$, let us assume that the bit error probability is $\varepsilon$, and let us indicate with $p_b = 1 - \varepsilon$ the corresponding bitwise decoding accuracy. By assuming that watermark bit errors are independent of each other, the probability of $k$ matches between $w$ and $\hat{w}$ is now:
\begin{equation}
	P(k) = \binom{n}{k} p_b^k (1-p_b)^{n-k},
	\label{eq:binomialH1}
\end{equation}
and the missed detection probability can be computed as:
\begin{equation}
	P_m = \sum_{k = 0}^{T-1} \binom{n}{k} p_b^k (1-p_b)^{n-k}.
	\label{eq:watermarked model}
\end{equation}
In Table~\ref{tab:P_m and k}, we list the minimum watermark accuracy $p_b$ required for different values of $P_f$ and $P_m$. It is clear from the table, that to get sufficiently low values of $P_f$ and $P_m$, the bitwise accuracy $p_b$ must be sufficiently large. For instance, in order to get $P_f = P_m = 10^{-4}$ ($n = 100$), it is necessary that $p_b \ge 0.83$.
\begin{table}[htbp]
	\caption{Minimum required $p_b$ for different values of $P_f$ and $P_m$ when $n=100$.
	}
	\label{tab:P_m and k} 
    \centering
    \begin{tabular}{c|cccccc}
    \toprule
	\diagbox{$P_f$}{$P_m$}& 0.05     & $10^{-2}$  & $10^{-3}$   & $10^{-4}$  & $10^{-5}$ & $10^{-6}$ \\ \midrule
	0.05 			& 0.60 & 0.65  & 0.69 & 0.72 & 0.75 & 0.77 \\
	$10^{-2}$       & 0.65   & 0.70  & 0.74 & 0.77 & 0.79 & 0.81\\
	$10^{-3}$   	& 0.70   & 0.74  & 0.78 & 0.81 & 0.83 & 0.85\\
	$10^{-4}$ 		& 0.73   & 0.77  & 0.81 & 0.83 & 0.85 & 0.87\\
	$10^{-5}$ 		& 0.75   & 0.80  & 0.83 & 0.86 & 0.87 & 0.89\\
	$10^{-6}$ 		& 0.78   & 0.83  & 0.86 & 0.88 & 0.89 & 0.90\\
    \bottomrule
    \end{tabular}
\end{table}

\section{Methodology}
\label{sec: THE PROPOSED METHOD}

In this section, we first introduce the concept of Wide Flat Minimum, and then we provide a thorough description of the watermark embedding process and the involved loss functions.

\subsection{What is a Wide Flat Minimum?}
\label{sec: What is wide flat minima?}
\begin{figure}[h]
	\centering
	\includegraphics[width=0.7\linewidth]{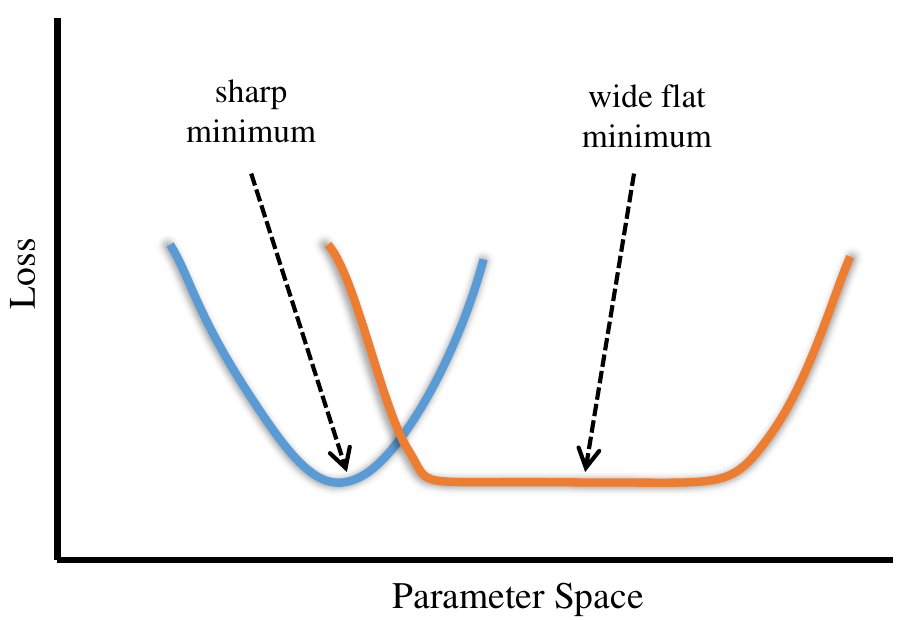}
	\caption{Comparison of a sharp minimum and a wide flat minimum in the one-dimensional case.}
	\label{fig:minima}
\end{figure}

A Wide Flat Minimum of a neural network refers to a large region in the parameter space where the loss value is small and approximately constant~\cite{hochreiter1997flat}.
Fig.~\ref{fig:minima} shows a toy example of a  flat wide minimum and a sharp minimum in a one-dimension case.
Requiring that training converges to a wide flat minimum has been proven to be helpful for the generalization capability of neural networks~\cite{Stutz2021,cha2021swad}. Recent works have proposed to look for a wide flat minimum for domain generalization~\cite{cha2021swad}, better robustness against adversarial attacks~\cite{Stutz2021}, and to fight catastrophic forgetting~\cite{shi2021overcoming}. In particular, our approach is inspired by~\cite{shi2021overcoming}, where the authors propose to alleviate catastrophic forgetting in continual learning by searching for a wide flat local minimum of the loss associated with the old task. They do so, by adding random noise to the model parameters during training and, jointly optimizing multiple instances of the losses, so that the loss values around the minimum are also small. When the model is fine-tuned to accomplish a new task, the model can efficiently learn the new task without forgetting the original task it had been trained on initially.

The design of a box-free GAN watermarking method which is robust against fine-tuning has some similarities with the necessity of avoiding catastrophic forgetting, given that fine-tuning the model should not imply forgetting the watermark embedded within the network during the initial training. In other words, the tasks addressed when the network is trained include image generation and watermark embedding ruled, respectively, by the adversarial and the watermark loss. During fine-tuning the network is trained on a new task by using only the adversarial loss. Given the flatness and width of the minimum of the watermark loss, the new task can be learned without {\em forgetting} the watermark.

\begin{figure*}
	\centering
	\includegraphics[width=0.93\linewidth]{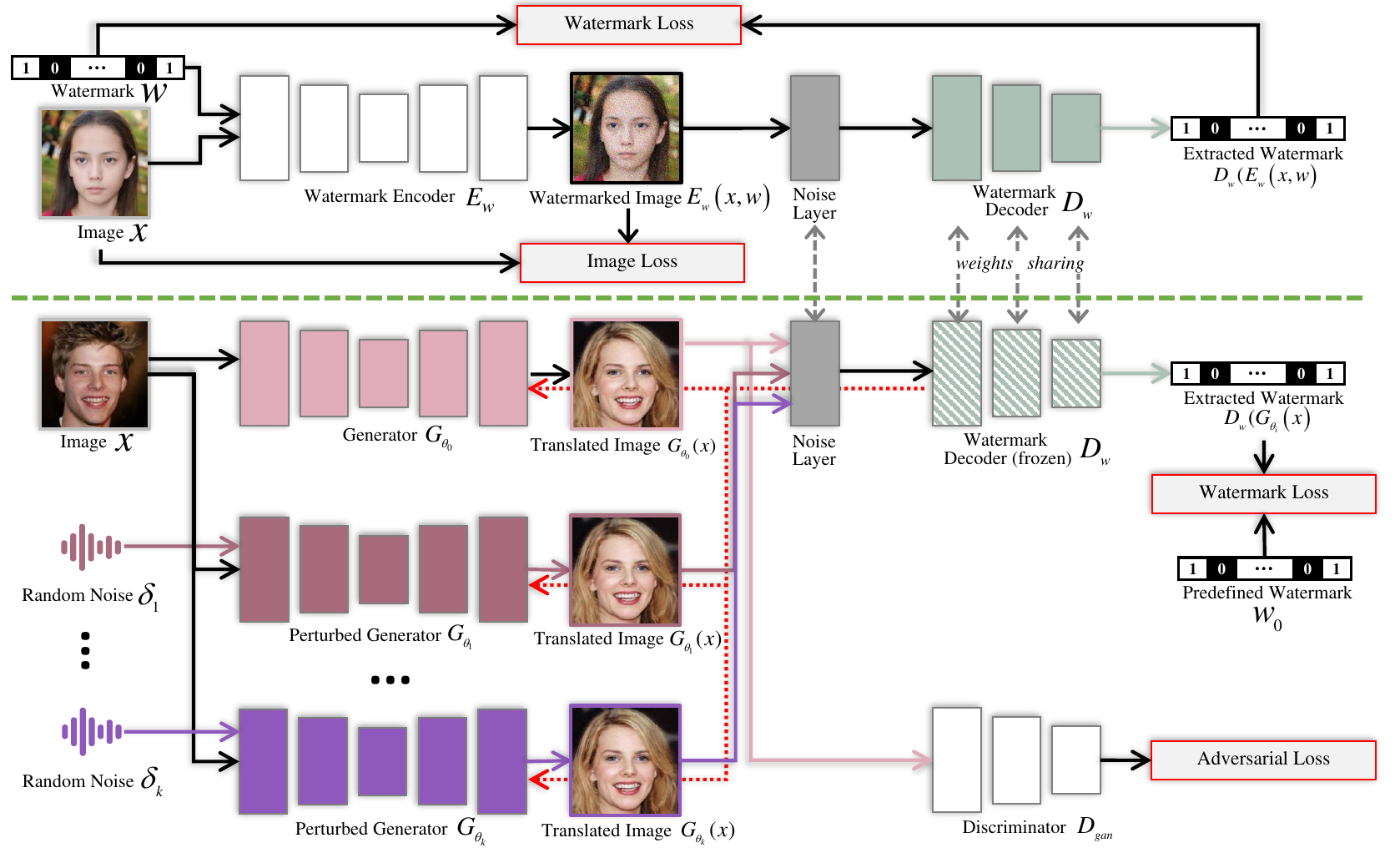}
	\caption{Overview of the proposed method using an image translation GAN (male$\rightarrow$female) as an example. Before watermarking the target GAN, we train an image watermarking network. Then, during GAN watermarking, the watermark decoder is frozen and used to extract the watermark bits from the generated images. The loss between a pre-defined watermark and the extracted watermark is back-propagated to instruct the generator to generate images containing the pre-defined watermark. A noise layer is introduced before the watermark decoder to enforce robustness against image-level attacks. For each iteration, we additionally sample $k$ random noise vectors and add them to the generator parameters, resulting in $k$ perturbed generators. The watermarking loss is calculated also on the perturbed generators, and the resulting gradients are used to update the generator. The red dashed lines indicate the backpropagation path of the watermarking loss from the frozen decoder to the generator.}
	\label{fig: overview}
\end{figure*}

\subsection{WFM watermarking}
\label{sec:Method Overview}

As shown in Fig.~\ref{fig: overview}, the WFM watermarking method proposed in this paper consists of two stages. In the first stage, we train a watermark encoder $E_w$ and a decoder $D_w$, which are jointly optimized by using an image reconstruction loss and a watermarking decoding loss:
\begin{equation}
    \mathcal{L}_w = \lambda_1 \text{MSE}(x,E_w(x,w))+ \lambda_2 \text{BCE}(w,D_w(E_w(x,w))),
    \label{eq: watermark loss}
\end{equation}
where MSE and BCE stand, respectively, for Mean Squared Error and Binary Cross Entropy, and $\lambda_1$ and $\lambda_2$ are weights balancing the two terms.
%
With regard to the second stage, let us consider a GAN whose goal is to learn a mapping function $G_\theta:X\rightarrow Y$ between domain $X$ and domain $Y$. $X$ may be a class of images belonging to a certain domain in image transfer applications, or a noise sample drawn from the uniform distribution, as it happens when the GAN is asked to generate images belonging to the output domain from scratch. In our method, the loss used to train $D_{gan}$ is not changed and can be expressed by:
\begin{equation}
	\begin{aligned}
		\mathcal{L}_{D_{gan}}  =& -  \mathbb{E}_{y \sim  Y} [\log D_{gan}(y)] \\ & -\mathbb{E}_{x \sim X}[\log (1-D_{gan}(G_\theta(x)))],
		\end{aligned}
		\label{eq:D loss}
\end{equation}
where $D_{gan}$ is asked to output 1 for original images and 0 for synthetic ones.
To let the generated images contain the owner watermark, say $w_o$, we modify the loss of $G_{\theta}$ as follows:
\begin{equation}
	\begin{aligned}
		\mathcal{L}_{G_{\theta}} & =  \mathbb{E}_{x \sim X}[\log (1-D_{gan}(G_{\theta}(x)))] \\
              & + \gamma \text{BCE}(w_o,D_w(G_{\theta}(x))),
    \end{aligned}
    \label{eq:gan loss}
\end{equation}
where $\gamma$ is a weight balancing the adversarial loss and the watermarking loss\footnote{To prevent falling into trivial solutions, $D_w$ is frozen in this stage.}. By optimizing $\mathcal{L}_{G_{\theta}}$ in Eq.~\eqref{eq:gan loss}, the generator learns to generate realistic images in the output domain containing the owner's watermark $w_o$\footnote{A similar architecture was used in~\cite{Fei2022}, here we modify it to achieve both image-level and model-level robustness.}.

To increase the robustness of the watermark against image-level processing, we include a noise layer between the generated images and the watermark decoder, which simulates image-processing attacks. In this way, the generator learns to embed a watermark that can be correctly retrieved even after several kinds of post-processing.

With regard to the robustness of the watermark against model-level manipulations (e.g., pruning, quantization, fine-tuning and surrogate model attacks), we force the training procedure to converge to a wide flat minimum of the watermarking loss term, so that any perturbation of the model within a certain range results in a loss value close to the minimum, with a negligible impact on the watermark decoding accuracy.
To this end, at each iteration, the watermarking loss term is computed on $k$ perturbed versions of the current parameters vector, namely $\theta_i = \theta_0 + \delta_i$, $i = 1 \dots k$, where $\delta_i$ is a perturbation vector taking values in the interval [$-b, b$], and $\theta_0$ is the parameters vector at the current iteration. Then the watermarking loss gradient used to update $\theta_0$ is computed on all parameter vectors, $\theta_0,\theta_1,...,\theta_k$, rather than on $\theta_0$, while the GAN loss is calculated as usual on $\theta_0$ only. Accordingly, the updating rule is given by
:
\begin{equation}
    \theta'_0=\theta_0 - \frac{\alpha}{k} \sum_{i=1}^{k}\nabla \mathcal{L}_w(\theta_i) - \beta \nabla \mathcal{L}_{G}(\theta_0),
\label{eq.updateruleWFM}
\end{equation}
where $\theta'_0$ is the updated parameters vector, $\nabla \mathcal{L}_w(\theta_i)$ is the gradient of the watermarking loss computed on $\theta_i = \theta_0 + \delta_i$, and $\alpha$ and $\beta$ are the weights that determine the relative importance of the watermarking and the GAN loss terms (see Algorithm~\ref{alg:algorithm1}). In this way, training converges to a parameter vector $\theta$, in whose surrounding the watermarking loss term assumes (approximately constant) small values.
\begin{algorithm}
	\caption{Search for a wide flat minimum of the watermarking loss.}
    \label{alg:algorithm1}
	\For{iteration $t=1,2,...$}{
		Store the initial generator parameters $\theta_0$\\
		\For{$i=1,2,...$}{
			Sample a noise vector $\delta_i$, s.t. $ -b< \delta_i < b$;\\
			Add $\delta_i$ to the parameters of the generators: $\theta_i = \theta_0 + \delta_i$;\\
			Compute $\mathcal{L}_w(\theta_i)=\text{BCE}(w_o, D_w(G_{\theta_i}(x)))$;\\
                Compute the gradient vector $\nabla \mathcal{L}_w(\theta_i)$;\\
	  	}
		Compute $\mathcal{L}_{G}(\theta_0)$ by Eq.~\eqref{eq:gan loss};\\
            Compute the gradient vector $\nabla \mathcal{L}_{G}(\theta_0)$;\\
            Updat $\theta_0$ as $\theta=\theta_0 - \frac{\alpha}{k} \sum_{i=1}^{k}\nabla \mathcal{L}_w(\theta_i) - \beta \nabla \mathcal{L}_{G}(\theta_0)$;
	}
\end{algorithm}
By optimizing the GAN with algorithm~\ref{alg:algorithm1}, the generator is encouraged to converge to a wide flat minimum of the watermarking loss,
with a radius approximately equal to $b$. If the watermarked generator is modified by pruning, quantization, or even fine-tuning, it maintains an acceptable watermark accuracy as long as the change in the weight vector is less than $b$.

\section{Implementation Details and Experimental Setting}
\label{sec:SETTINGS}

\begin{table}[htbp]
	\caption{Summary of task and datasets}
	\label{tab:Summary of task and datasets}
    \centering
    \begin{tabular}{c|ccccc}
    \toprule
    Tasks   & Architecture        & Dataset     & Num. of Images        \\ \midrule
    Image Generation  & StyleGAN2    & FFHQ  &  70,000 \\
    Sytle transfer    &   CycleGAN     &  Photo2Monet  &   8,231     \\
    Image Synthesis  & SPADE    & ADE20K  &  22,210   \\
    Image Editing    & StarGAN v2   & CelebA-HQ  &  30,000    \\
     Super Resolution &  CFSRCNN    &  VOC2012 &   1,7125    \\
     Denosing &  MPRNet    &  SIDD &   30,000   \\

    \bottomrule
    \end{tabular}
\end{table}

\subsection{Models and Datasets}
The proposed GAN watermarking is both model- and task-agnostic, thus, we evaluated its performance on 6 different tasks and architectures.

\paragraph{Image Generation} This task aims at generating fully synthetic images starting from a random noise input.
We used StyleGAN2~\cite{Karras2020} as target model, and FFHQ~\cite{Karras2019} as training dataset to generate facial images.

\paragraph{Sytle transfer} This task aims at transferring the style of the source image to a target style.
We used CycleGAN~\cite{zhu2017unpaired} as target model, and Photo2Monet as training dataset.
CycleGAN is trained to translate natural photos into Monet-style paintings.

\paragraph{Image Synthesis} This task aims at synthesizing realistic images using semantic layouts.
We used SPADE~\cite{Park2019} as target model and ADE20K~\cite{Zhou2017} as training dataset.

\paragraph{Image Editing} This task aims at editing the attributes of a given image, for example, changing the hair color or gender of a human face.
We used StarGAN v2~\cite{Choi2020} as target model, and CelebA-HQ~\cite{Karras2017} as training dataset.
The model is trained to edit the face attributes of the input images.

\paragraph{Super Resolution} This task aims at upsampling low-resolution images to high-resolution images.
We used CFSRCNN~\cite{tian2020coarse}, a fully convolutional network as target model, and VOC2012~\cite{Everingham2015} as training dataset.
CFSRCNN is trained to scale up images by a factor of 4.

\paragraph{Denosing} This task aims at removing noise from the input images.
We used MPRNet~\cite{zamir2021multi}, a U-Net based network as target model, and SIDD~\cite{abdelhamed2018high} as training dataset.

Table~\ref{tab:Summary of task and datasets} summarizes the 6 different tasks evaluated in this paper and provides information about the corresponding datasets.
We chose these different models to prove the generality of the proposed method.
Please note that the general description given in Section~\ref{sec: THE PROPOSED METHOD} takes a GAN architecture as baseline, however, our approach also works with other kinds of networks.
For some tasks, in addition to the native adversarial loss, there are other losses involved, such as the cycle loss and identity loss in CycleGAN~\cite{zhu2017unpaired}.
For simplicity, we omitted these losses in Section~\ref{sec: THE PROPOSED METHOD}, since this omission does not affect the watermark embedding part of the process. In the sequel, for simplicity, we indicate the models trained for the various tasks by referring to the architecture.

\subsection{Evaluation Metrics}
To evaluate the effectiveness of the proposed approach, we used the following metrics:

\paragraph{Bitwise Accuracy (Bit Acc)} Bit Acc, also indicated by $p_b$, measures the fraction of correctly predicted bits. A high bitwise accuracy is preferable. The minimum $p_b$ required for accurate verification can be set by using the analysis in Section~\ref{subsection: Ownership Verification} and setting desired values of $P_f$ and $P_m$.

\paragraph{Fréchet Inception Distance (FID)} FID~\cite{Heusel2017} is a widely used metric to evaluate the quality of the images generated by a GAN. It measures the distance between the distribution of real samples and generated samples in the feature space. A low FID is preferable.

\paragraph{Peak Signal-to-Noise Rate (PSNR)} PSNR measures the pixel-wise difference between the generated and target images produced by the super-resolution and denoising tasks. A high PSNR is desirable.

\subsection{Implementation Details}
\label{sec:Implementation Details}

All experiments have been carried out by using PyTorch 1.9 on Centos 7.0, with RTX 3090 GPU. The models used for the comparisons have been trained by using the official code, and the optimizer, learning rate, batch size, and number of training epochs are the same. All models have been initially trained without the watermarking loss term until convergence and then fine-tuned with both the original loss and the watermarking loss to embed the watermark. For the noise layer, we randomly applied addition of Gaussian noise with standard deviation in [0.0001, 0.10], Gaussian blurring with kernel sizes in [0, 5] and standard deviation in [1, 7], differentiable JPEG compression as defined in~\cite{Zhu2018}, and color transformations including brightness, contrast, and saturation adjustment in the range [0.0, 0.05]. Each processing operator is applied with probability 0.15. The watermark decoder is frozen during embedding. The watermark consists of 100 bits, the weights $\lambda_1$ and $\lambda_2$ used to train the watermark encoder and decoder are both 1.0. In the embedding stage, the number of random noise perturbations $k$ is 4, and the weight of the watermark loss $\gamma$ is 2.0. As to $b$, we used 0.03 for all models but StyleGAN2 for which we let $b = 1.0$. Eventually, we let $\alpha = \frac{1}{4}$ and $\beta = 1.0$.

\section{Experimental results}
\label{sec: EXPERIMENTS}

\subsection{Effectiveness}
In Table~\ref{tab:Bit Acc of different models}, we report the bitwise accuracy of WFM watermarking on four different image generation tasks (see Table \ref{tab:Summary of task and datasets}). The first column indicates the watermarked model, while the second indicates the watermarking algorithm. Specifically, we compared the performance of our method with those described in~\cite{Yu2021} and~\cite{Fei2022}. Both are multi-bit box-free GAN watermarking methods and~\cite{Fei2022} can be considered as the baseline of the proposed method.

We can observe that the proposed method achieves nearly perfect accuracy, except for the SPADE model.
For image generation models such as StyleGAN2 and StarGAN v2, we have achieved 0.999 Bit Acc ($p_b$). Although~\cite{Yu2021} is also effective on multiple tasks, our method has a significant advantage in terms of Bit Acc. Meanwhile, our method also slightly outperforms~\cite{Fei2022} on 3 out of 4 tasks. Table~\ref{tab:Bit Acc of different models} also reports the missed detection probability at $P_f = 10^{-3}$ estimated by using the analysis described in Section~\ref{subsection: Ownership Verification}. In all cases, the missed detection probabilities $P_m$ are very close to zero.

\begin{table}[htbp]
	\renewcommand{\arraystretch}{1.1}
	\caption{Bit Acc ($p_b$) of different GAN watermarking methods over different GANs and corresponding $P_m$ (computed as in Eq.~\eqref{eq:watermarked model} ) when the detection threshold is set in such a way to get $P_f = 10^{-3}$.
	}
	\label{tab:Bit Acc of different models}
    \centering
    \begin{tabular}{m{1.5cm}<{\centering}|m{1.5cm}<{\centering}|m{1.3cm}<{\centering}m{2cm}<{\centering}}
    \toprule
	Models                              & Methods     & Bit Acc ($p_b$)   & $P_m$  \\ \midrule
    \multirow{3}{*}{StyleGAN2}  &    Yu~\textit{et al.}~\cite{Yu2021} &0.854  & $5.9 \times 10^{-07}$\\
                                & Fei~\textit{et al.}~\cite{Fei2022}   &0.999  & $1.5 \times 10^{-75}$ \\
                                &  Ours                 &0.999  & $1.4 \times 10^{-102}$ \\\midrule

    \multirow{3}{*}{CycleGAN}  & Yu~\textit{et al.}~\cite{Yu2021} &0.958   &      $5.1 \times 10^{-20}$ \\
                                & Fei~\textit{et al.}~\cite{Fei2022}   &0.996&     $9.4 \times 10^{-51}$\\
                                &  Ours                  &0.998&        $6.5 \times 10^{-58}$\\\midrule

    \multirow{3}{*}{SPADE}  &    Yu~\textit{et al.}~\cite{Yu2021}     &0.979  &$9.2 \times 10^{-29}$ \\
                                & Fei~\textit{et al.}~\cite{Fei2022}   &0.989  &$1.8 \times 10^{-37}$\\
                                &  Ours                 &0.968   &$1.7 \times 10^{-23}$\\\midrule

    \multirow{3}{*}{StarGANv2}  &    Yu~\textit{et al.}~\cite{Yu2021} &0.941 & $8.9 \times 10^{-17}$\\
                                & Fei~\textit{et al.}~\cite{Fei2022}   &0.999 & $3.2 \times 10^{-80}$\\
                                &  Ours                 &0.999  & $1.4 \times 10^{-102}$\\
    \bottomrule
    \end{tabular}
\end{table}

\subsection{Fidelity}

Fidelity is a critical requirement for an effective watermarking method since it measures the impact that the presence of the watermark has on the performance of the model on its primary task. For image generation tasks, fidelity is measured by the perceptual quality of the generated images.

Table~\ref{tab:FID of different models} shows the FID of the baseline model, i.e., the non-watermarked model, and the FID of the model watermarked by different methods. Note that the estimation of FID is influenced by the number of images, and an insufficient number of images can lead to inaccurate FID. Typically, the recommended number of images for FID calculation is 50,000. However, the number of images in the validation datasets for CycleGAN and SPADE is much less than 50,000, so we report the FID of CycleGAN and SPADE for reference only. Still, we can observe that, over four different tasks, our method has a very slight effect on FID with almost no deterioration in image quality.
%
For StyleGAN2 and StarGAN v2, our method only increases the FID by 0.08 and 0.10, which is acceptable. For CycleGAN and SPADE, the initial FID is rather high, however, our method only increases it to a value  2 / 3 points larger. The method in~\cite{Yu2021} has a much stronger impact on
quality of the images generated by the GAN.

\begin{table}[htbp]
	\renewcommand{\arraystretch}{1.1}
	\setlength\tabcolsep{2pt}
	\caption{FID of different models before and after watermark embedding with different methods.}
	\label{tab:FID of different models}
    \centering
    \begin{tabular}{m{2cm}<{\centering}|m{1.5cm}<{\centering}m{1.5cm}<{\centering}m{1.5cm}<{\centering}m{1.5cm}<{\centering}}
    \toprule
	Methods$\textbackslash$ Models  & StyleGAN2    &CycleGAN & SPADE & StarGAN v2 \\  \midrule

	Baseline 		    &5.28	  &	47.68  & 46.80    &   13.81\\
    Yu~\textit{et al.}~\cite{Yu2021}  	& 6.61  & 50.34   & 59.42   &   14.89                  \\
    Fei~\textit{et al.}~\cite{Fei2022}  &5.16	& 49.39   & 50.70      &13.53\\
    Ours  			    &5.36   & 49.52    &  50.59    &13.91  \\
    \bottomrule
    \end{tabular}
\end{table}

\begin{figure*}
	\centering
	\includegraphics[width=\linewidth]{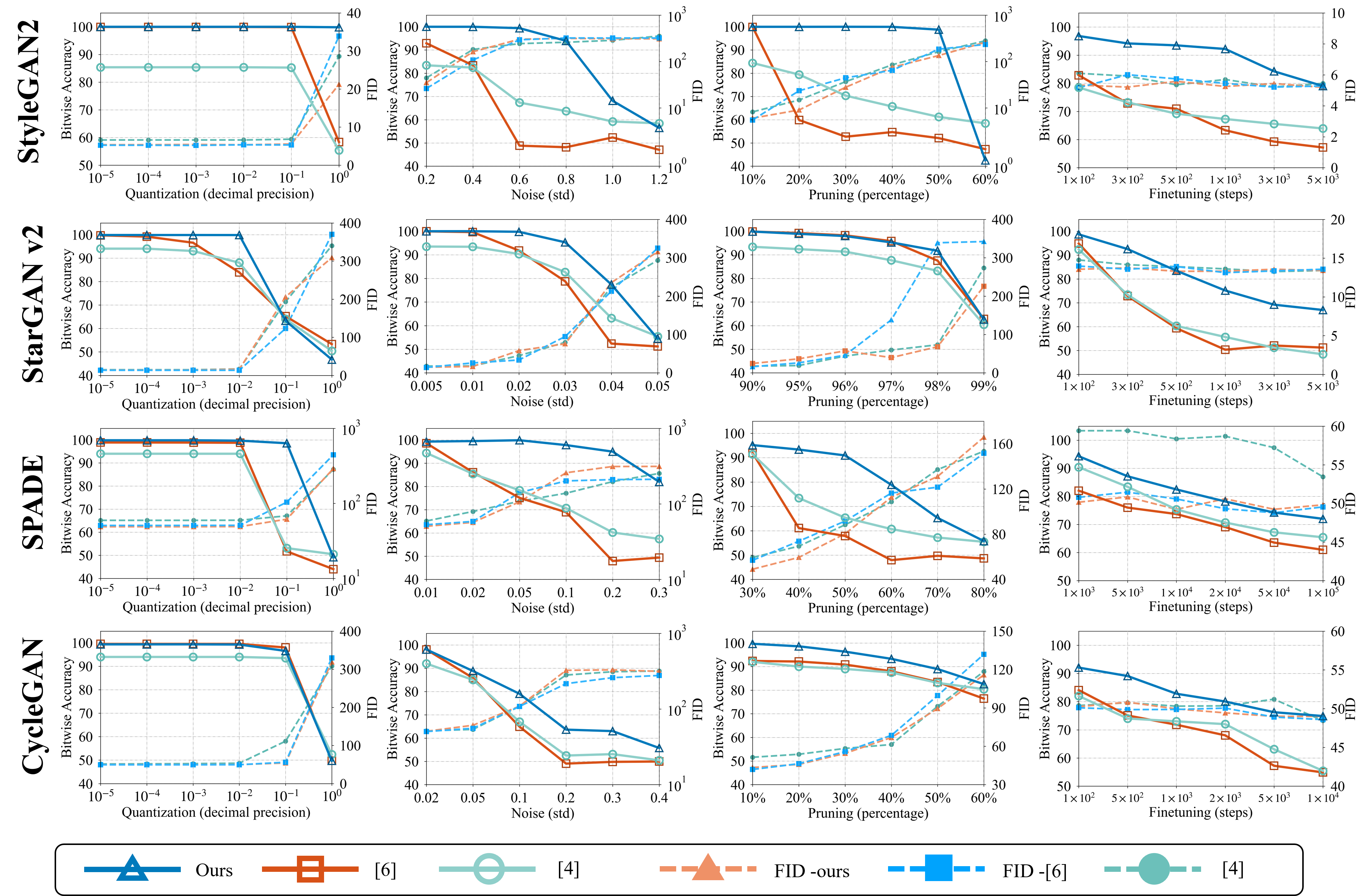}
	\caption{Robustness to different model-level attacks.}
	\label{fig:robustness_to_model_attacks}
\end{figure*}

\subsection{Robustness to model-level attacks}

We evaluated the robustness of our methods against pruning, quantization, noise addition, and fine-tuning. For pruning, we set the weights with the smallest absolute values to zero. For quantization, we represented the model weights with a given decimal precision. For noise addition, we added Gaussian noise to the model weights. For fine-tuning, we fine-tuned the generator with the original training set, but without using the watermarking loss term.

Fig.~\ref{fig:robustness_to_model_attacks} reports (left y-axis of the plots) the Bit Acc $p_b$ obtained after the models are modified by quantization, noise addition, pruning and fine-tuning. The FID obtained with the attacked models is also reported on the right y-axis\footnote{For sake of visualization, a different x-axis range is used in the various cases for the same attack.}
Compared to both~\cite{Fei2022} and~\cite{Yu2021}, our solution significantly improves the robustness of the watermark against all model-level attacks. The WFM has the most beneficial effect on StyleGAN2 (image generation). The Bit Acc ($p_b$) remains almost 1 even though the model weights are compressed by decimal precision $10^0$, meaning that the model weights are represented by integer numbers.
In~\cite{Fei2022} the Bit Acc $p_b$ drops to about 0.50 after the addition of Gaussian noise with standard deviation (std) 0.6, or 50\% of the parameters are pruned, while our solution can effectively resist this noise. In the fine-tuning attack, our method is able to maintain a $p_b$ over 0.80 after about 5 epochs (5,000 steps) fine-tuning on FFHQ~\cite{Karras2019},
while it takes only about 100 steps to remove the watermark embedded as in~\cite{Fei2022} and~\cite{Yu2021}.

In the case of StarGAN~v2, although the benefit against quantization, noise addition and pruning attacks is not readily apparent, robustness to fine-tuning is much improved with our method and the Bit Acc $p_b$ is at least 0.20 higher w.r.t. ~\cite{Fei2022} and~\cite{Yu2021}.
For SPADE, the Bit Acc $p_b$ of our method is approximately at least 0.20 to 0.30 better than~\cite{Fei2022} before the model is completely disabled by noise addition and pruning attacks.

In summary, our method yields a higher Bit Acc than~\cite{Fei2022} and~\cite{Yu2021} under model-level attacks with the same strength, demonstrating the effectiveness of our method. It is worth noting that the adversary may be willing to remove the watermark at the cost of degrading the image quality to a certain extent. For example, by pruning 20\% of the weights of StyleGAN2, the FID is increased to about 10, and the Bit Acc of~\cite{Fei2022} is drastically reduced to 0.60, the Bit Acc of~\cite{Yu2021} is decreased to 0.80, while the Bit Acc of our method remains unchanged. In fact, to remove the watermark embedded using our method, the attacker has to modify the network down to a point of making it unusable, e.g., the FID must be raised to 300. Note that according to the analysis in Sect.~\ref{subsection: Ownership Verification} $p_b = 0.8$ is enough to achieve ownership verification with $P_f$ and $P_m$ in the order of $10^{-4}$ to $10^{-3}$.

\begin{figure*}[ht]
	\centering
	\includegraphics[width=\linewidth]{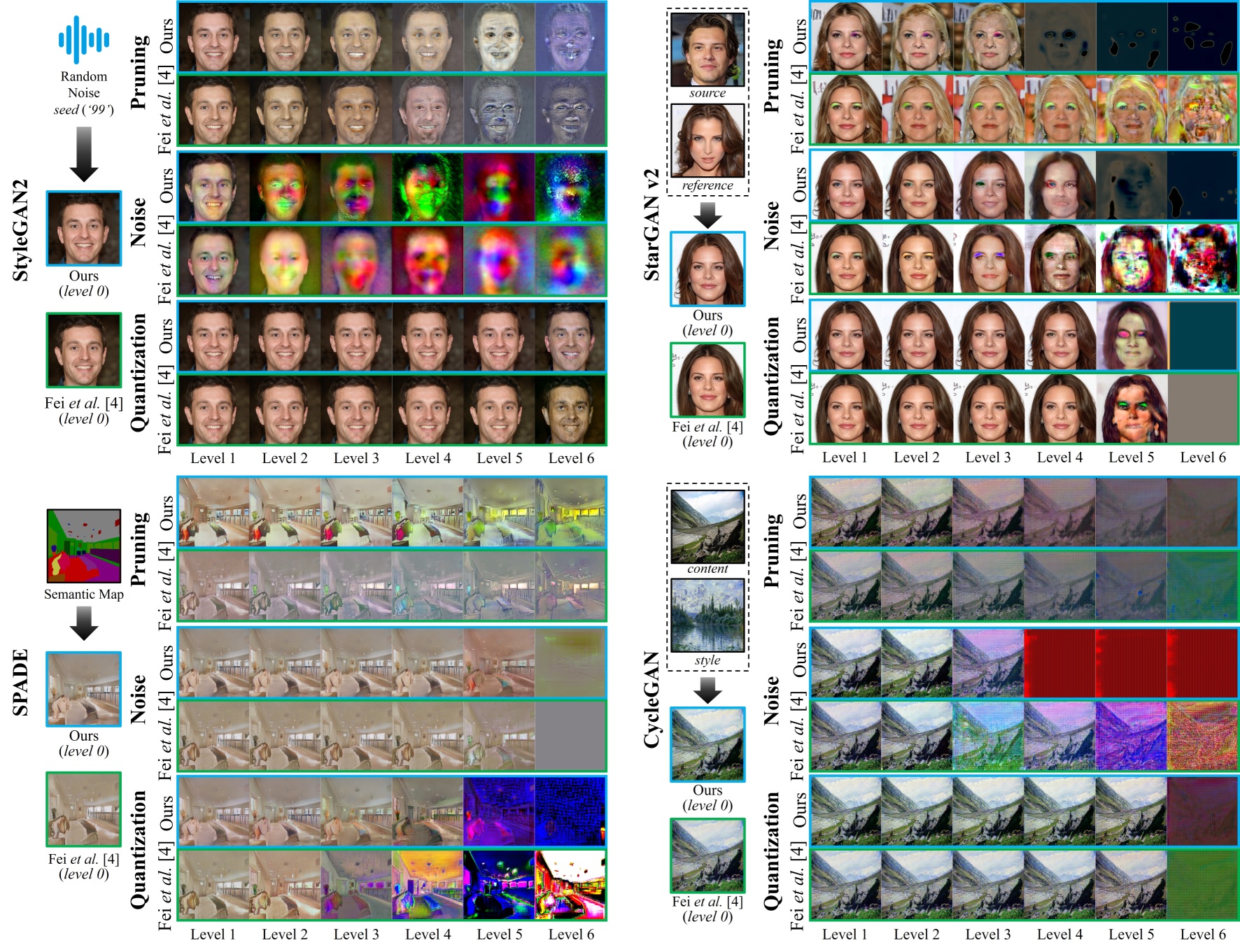}
	\caption{Samples generated by watermarked models after the attacks (quantization, noise, pruning) applied with various strengths.}
	\label{fig:samples_of_model_attacks}
\end{figure*}

Fig.~\ref{fig:samples_of_model_attacks} shows some images generated by the attacked models under different kinds of attacks and intensities. For simplicity, we denote the strength by levels 1 to 6, which correspond to the ticks on the x-axis in Fig.~\ref{fig:robustness_to_model_attacks}. We can observe that in many cases the watermark accuracy is sufficient to achieve ownership authentication despite severe image degradation (e.g., in Level 3 and 4). Therefore, removing the watermark goes at the cost of poor quality of the generated images, making the models virtually useless.

\begin{figure*}[ht]
	\includegraphics[width=0.95\linewidth]{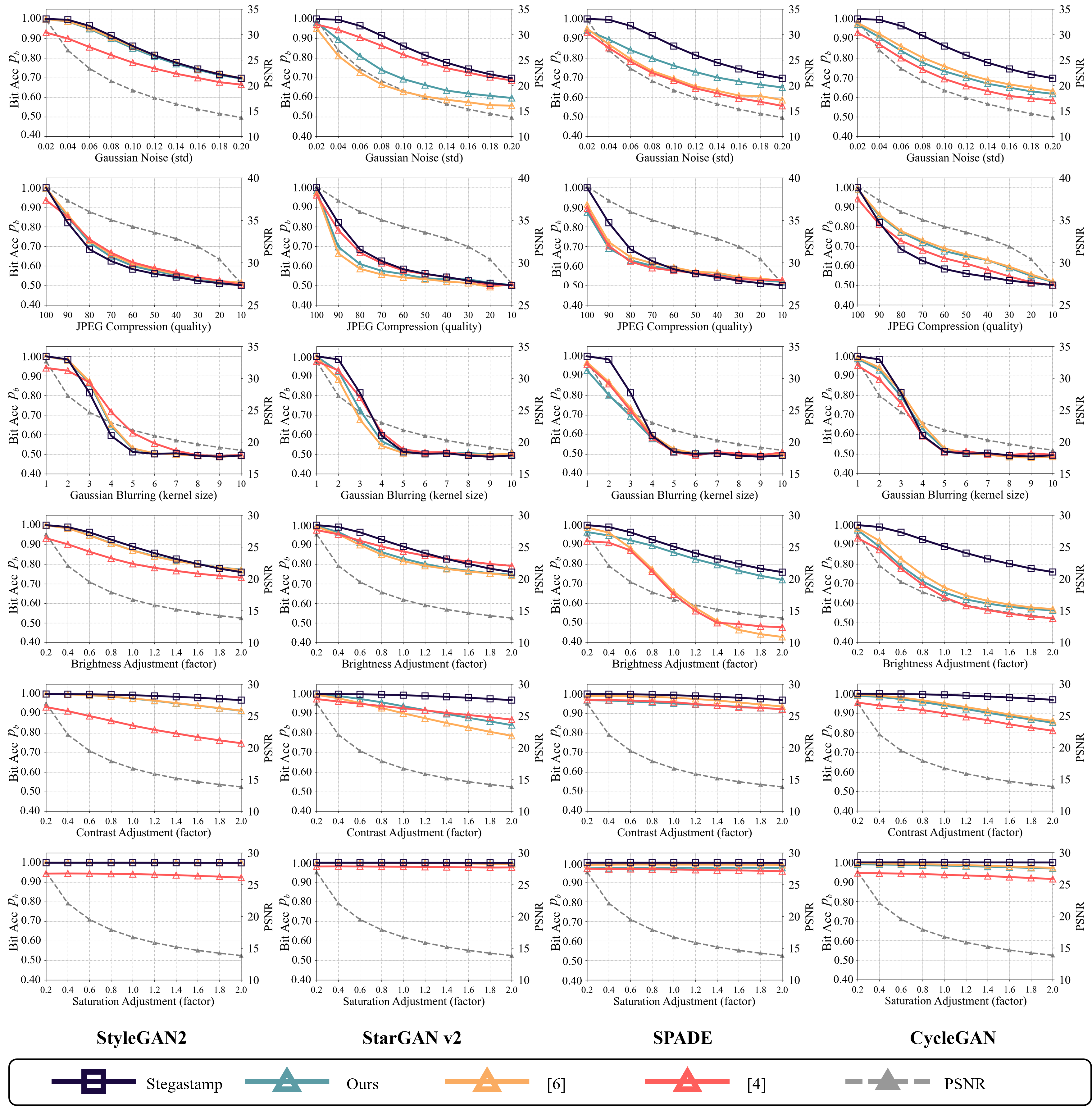}
	\caption{Robustness to different image-level attacks. The gray dashed lines denote the PSNR between the original image and the processed image.}
	\label{fig:robustness_to_image_attacks}
\end{figure*}

\subsection{Robustness to surrogate model attacks}

In the previous section, we considered attacks carried out in white-box conditions, where the model can be directly accessed by the attacker. In this section, we consider a scenario where the attacker can only query the model and apply a surrogate model attack. We carried out surrogate model attacks on the watermarked CycleGAN with the same and different architectures of the attacked model. The two surrogate models are based on ResNet (the same as the attacked model) and U-Net (different from the attacked model) architectures. Both surrogate models are trained for 200 epochs by using input and output pairs generated by the watermarked CycleGAN generator.
We also evaluated the effect of training the surrogate models with different loss functions. Specifically, we optimized the surrogate models by using: i) MSE loss $\mathcal{L}_{2}$ only, and ii) MSE loss $\mathcal{L}_{2}$ and adversarial loss $\mathcal{L}_{adv}$. In the latter case, we used the discriminator of the original CycleGAN and set the weight of the adversarial loss to 0.001. The results we got are shown in Table~\ref{tab:surrogate}.

\begin{table}[htbp]
	\renewcommand{\arraystretch}{1.1}
	\setlength\tabcolsep{2pt}
	\caption{Bit Acc $p_b$ after surrogate model attacks with different losses and network architectures.}
	\label{tab:surrogate}
    \centering
    \begin{tabular}{m{2cm}<{\centering}|m{2cm}<{\centering}m{2cm}<{\centering}m{2cm}<{\centering}}
    \toprule
	Method   &  Loss  & ResNet-based    &  UNet-based  \\  \midrule

	\multirow{2}{*}{Fei \textit{et al.} \cite{Fei2022}} &$\mathcal{L}_{2}$   &0.727	  &	0.704  \\
    &$\mathcal{L}_{2} + \mathcal{L}_{adv}$  	& 0.662  & 0.669   \\ \midrule
	\multirow{2}{*}{Yu \textit{et al.} \cite{Yu2021}}&$\mathcal{L}_{2}$    &0.735	  &	0.736  \\
    &$\mathcal{L}_{2} + \mathcal{L}_{adv}$   & 0.686  & 0.660\   \\
	\midrule
	\multirow{2}{*}{Ours}&$\mathcal{L}_{2}$	       & 0.793	  &	0.786  \\
    &$\mathcal{L}_{2} + \mathcal{L}_{adv}$  	& 0.722  & 0.725   \\

    \bottomrule
    \end{tabular}
\end{table}

We can observe that our method achieves nearly 0.80 Bit Acc when the surrogate model adopts the same architecture of the attacked model and is trained by using an $\mathcal{L}_{2}$ loss, which is 0.06 $\sim$ 0.07 higher than the $p_b$ obtained by the other methods. This confirms that the watermark can be transferred to the new network.
We also observe that the architecture used to build the surrogate model has only a slight influence on the watermark accuracy. Even when we used a completely different architecture, there is no more than a 0.01 drop in $p_b$ for our method.
When the surrogate model is trained by using both $\mathcal{L}_{2}$ and $\mathcal{L}_{adv}$ losses, the watermark accuracy decreases by 0.06 $\sim$ 0.07.
However, watermark accuracies of both surrogate models are still above 0.70, noticeably higher than the accuracies obtained with the other methods. Moreover, as we have proven in Section~\ref{subsection: Ownership Verification}, when the Bit Acc $p_b$ is larger than 0.70 both $P_f$ and $P_m$ are less than $10^{-2}$.

\subsection{Robustness to image processing manipulations}
\label{sec: Robustness to image processing attacks}

We also evaluated the robustness of our method against the image processing operations included in the noise layer used during training, see Fig.~\ref{fig:robustness_to_image_attacks}.
The figure also reports the results obtained by~\cite{Fei2022} and~\cite{Yu2021},
which have been trained by using the same noise layers. The left y-axis indicates the watermark extraction accuracy on the processed images, while the right y-axis indicates the PSNR between the original images and the processed images. In addition to the results of different GAN watermarking methods, we also plotted the results of Stegastamp~\cite{Tancik2020} as baseline, namely the image watermarking network on which the watermark decoder considered here and in~\cite{Fei2022} are based.
Since the generator is not trained together with the watermark decoder, the robustness of all the GAN watermarking methods are expected to be worse than this baseline.

\begin{figure}[ht]
	\includegraphics[width=0.9\linewidth]{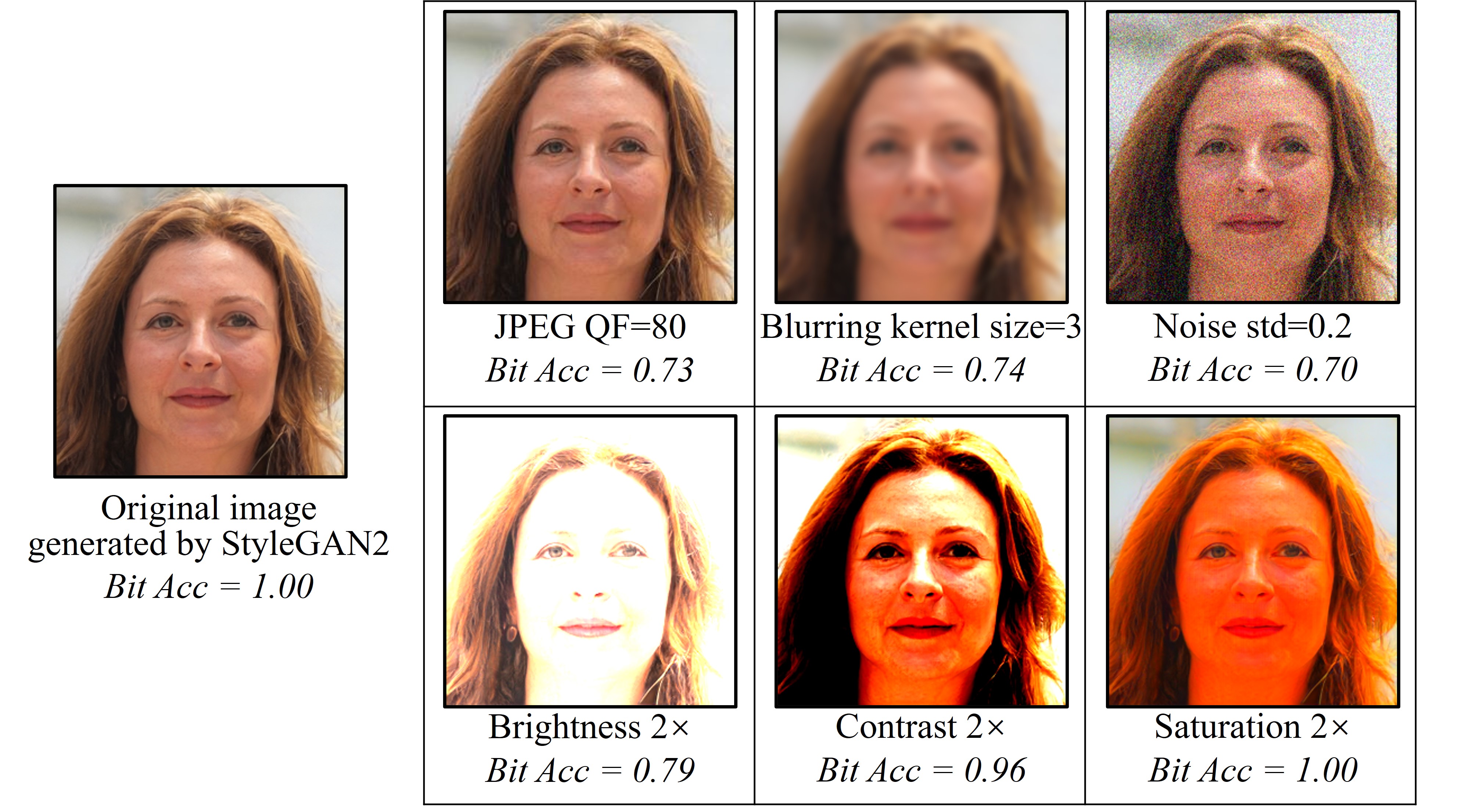}
	\caption{Samples processed by various operations and Bit Acc.}
	\label{fig:model_attacks_total_samples}
\end{figure}

In general, with the increase of the image processing intensity, the Bit Acc of all methods decreases, while the PSNR also drops significantly. We can observe that in most cases, the original Stegastamp achieves the best robustness, being this the upper limit of GAN watermarking methods.
All methods demonstrate promising robustness against color adjustment, while JPEG compression and Gaussian blurring are more challenging, and the Bit Acc decreases as the intensity of the attack increases.
Nonetheless, watermark removal requires a compromise in terms of image quality (PSNR). In Fig.~\ref{fig:model_attacks_total_samples}, we show some samples of StyleGAN2 images after the application of processing with the intensity such that the Bit Acc $p_b$ remains above 0.70 (corresponding to a $P_f$ and $P_m$ lower than $10^{-2}$ for the ownership verification). We can  argue that all attacks need to be very strong to reduce the Bit Acc to a value that makes the ownership verification fail, except for JPEG compression, which leaves only some artifacts in the image.

\subsection{Effectiveness on image processing models}

In the experiments reported in the previous sections, all watermarked models are based on GANs, although for different image generation or synthesis tasks.
To demonstrate that our method can be applied to completely different generative architectures, in this section, we present the results we got by watermarking 2 CNN-based image processing networks, namely MPRNet~\cite{zamir2021multi} and CFSRCNN~\cite{tian2020coarse}.
Note that both models are optimized by using pixel-wise MSE, without adversarial loss.

We report the results of Bit Acc and PSNR in Table~\ref{tab:Bit acc for image processing models}. We can observe that both our method and~\cite{Fei2022} achieve nearly perfect accuracy. In contrast, for both MPRNet and CFSRCNN, the method in~\cite{Yu2021} cannot converge. In terms of fidelity, the impact of our method on the original task is within an acceptable range. In fact, our watermarking method reduces the PSNR by no more than 0.5dB on MPRNet and about 2dB on CFSRCNN. Compared to~\cite{Fei2022}, we achieve a better robustness with a very low impact on the performance of the original task.

\begin{figure}[ht]
	\includegraphics[width=\linewidth]{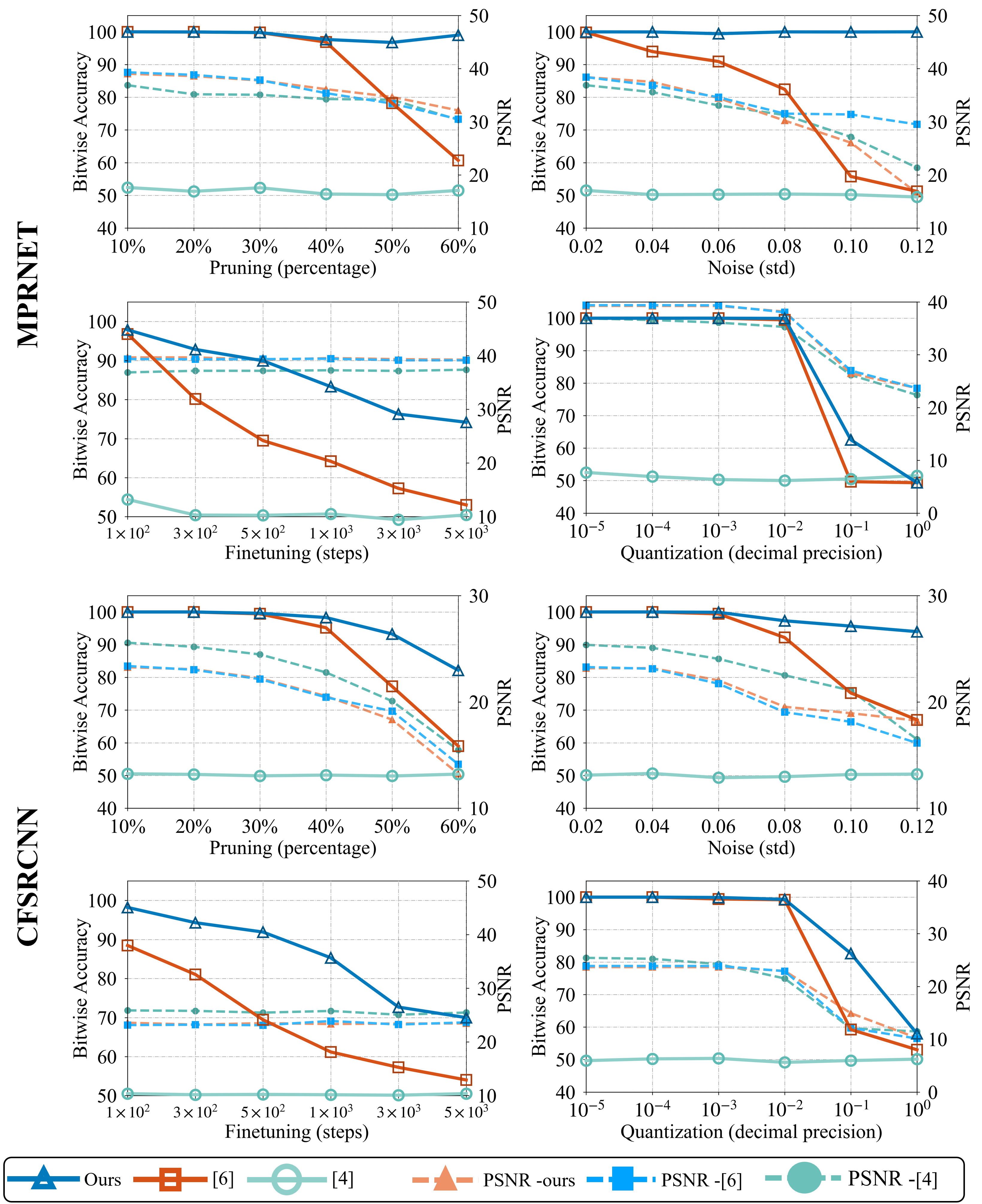}
	\caption{Robustness of CNN-based image processing networks against model-level attacks.}
	\label{fig:robustness_to_model_attacks2}
\end{figure}

Fig.~\ref{fig:robustness_to_model_attacks2} shows the robustness against model-level attacks for MPRNet and CFSRCNN.
%
%
%
Our method shows the same strong robustness for both networks. When 60\% of the network parameters are set to zero, $p_b$ for MPRNet still remains nearly 1.00, and over 0.80 for CFSRCNN, while it drops to 0.60 for~\cite{Fei2022}.
For the most challenging fine-tuning attack, our method is able to maintain an accuracy of 0.70, while~\cite{Fei2022} is reduced to about 0.50.
Overall, we observe that in the case of CNN-based image processing networks, the watermark is less robust than for GANs.
For instance, compared to the results in Fig.~\ref{fig:robustness_to_model_attacks}, the reduction in Bit Acc is more significant under Gaussian noise addition of the same intensity.
However, a   gain in robustness is always  achieved with our method 
with respect to~\cite{Yu2021} and~\cite{Fei2022} also in this case of  image processing networks. This gain is a significant one in many cases, especially against pruning and fine-tuning attacks.
\begin{table}[htbp]
	\renewcommand{\arraystretch}{1.1}
	\setlength\tabcolsep{3pt}
	\caption{Bit Acc ($p_b$), $P_m$ (when $P_f = 10^{-3}$), and PSNR for image processing models.}
	\label{tab:Bit acc for image processing models}
    \centering
    \begin{tabular}{m{1.5cm}<{\centering}|m{2.5cm}<{\centering}|m{1.0cm}<{\centering}m{1.9cm}<{\centering}m{0.8cm}<{\centering}}
    \toprule
     Task                                &   	Model  & Bit Acc   & $P_m$ & PSNR  \\ \midrule
    \multirow{4}{*}{Denosing}    &    	MPRNet -baseline 	& -   &  - &   39.70\\
                                      &    MPRNet -\cite{Yu2021} 	& 0.565  & $\approx0.90$  &   39.64\\
                                  &    MPRNet -\cite{Fei2022} 	& 0.999   &  $\approx1.2 \times 10^{-96}$    & 39.39  \\
                                &  MPRNet -ours 	& 0.999       &  $\approx1.3 \times 10^{-102}$  & 39.24\\   \midrule
    \multirow{4}{*}{SR}    &    	CFSRCNN -baseline 	& -   &  - & 25.92\\
                                       &   CFSRCNN -\cite{Yu2021} 	& 0.511  &     $\approx0.99$       &   25.92\\
                                   &   CFSRCNN -\cite{Fei2022} 	& 1.000  &     $\approx0$       &   23.86\\
                                 & CFSRCNN -ours 	& 1.000  &     $\approx0$    &23.61\\
    \bottomrule
    \end{tabular}
\end{table}

\subsection{Ablation study}
In this section, we study the effect of the number $k$ of samples used within each training step to converge to a wide flat minimum and the noise range $b$ (see Algorithm 1) on the fidelity of the model, and on the accuracy and robustness of the watermark.
Intuitively speaking, a larger number of $k$ could potentially explore more directions in the parameters space, and force the model to a flatter minimum, while a larger noise range could broaden the scope of the minimum.
To verify this, we took StyleGAN2 as a case study and watermarked it using different values of $k$ and $b$. The resulting watermark Bit Acc and FID are shown in Table~\ref{tab:k and b}.
We can observe that, for a given $b$, as $k$ increases, the Bit Acc remains stable and does not increase further.
However, when $k$ is fixed and $b$ increases, the Bit Acc tends to reduce, and the FID shows a clear upward trend.
These results suggest that the parameter space is sufficiently high-dimensional and has sufficient redundancy, allowing to converge to a minimum that can accommodate a moderate perturbation.
Nonetheless, as the perturbation magnitude $b$ increases, converging to such a minimum while ensuring the quality of the generated images is highly challenging, and such a minimum may not even exist.

\begin{table}[htbp]
	\caption{Bit Acc $p_b$ and FID under different noise sampling iterations $k$ and noise ranges $b$ for StyleGAN2.}
	\label{tab:k and b}
    \centering
    \begin{tabular}{c| cccccc}
    \toprule
	\diagbox{$k$}{$b$} & 0.1   & 0.3   & 1.0  & 3.0  & 5.0  & 10.0 \\ \midrule
	 1 				   & \makecell{0.999 \\ 5.48} & \makecell{0.997 \\5.37} & \makecell{0.999 \\5.38}  & \makecell{0.973 \\ 6.17} & \makecell{0.965 \\7.04}  & \makecell{0.944 \\ 9.13} \\ \midrule
	
	 4 		           & \makecell{0.999 \\ 5.28}  & \makecell{0.999  \\ 5.48}	 & \makecell{0.999 \\5.36} & \makecell{0.971  \\5.51} & \makecell{0.957 \\8.10}	 & \makecell{0.945  \\ 11.60 } \\ \midrule

	 8 		           & \makecell{0.996 \\ 5.41}   & \makecell{0.996 \\ 5.91}	 & \makecell{0.993 \\ 5.67} & \makecell{0.963 \\ 6.51} &	 \makecell{0.957 \\ 7.69}	& \makecell{0.948\\ 12.10}  \\ \midrule
	
	 16 		       & \makecell{0.998 \\5.51}  & \makecell{0.996 \\ 5.87 }	& \makecell{0.996 \\6.17}	 & \makecell{0.965 \\7.09	}	 & \makecell{0.950 \\8.85}  & \makecell{0.944 \\9.85}	\\

    \bottomrule
    \end{tabular}
\end{table}

\begin{table}[htbp]
	\caption{Bit Acc $p_b$ after fine-tuning attack under different noise sampling times $k$ and noise ranges $b$ for StyleGAN2.}
	\label{tab: ft k and b}
    \centering
    \begin{tabular}{c| cccccc}
    \toprule
	\diagbox{$k$}{$b$} & 0.1   & 0.3   & 1.0  & 3.0  & 5.0  & 10.0 \\ \midrule
	 1 				   & 0.627 & 0.699 & 0.722 & 0.644 & 0.591 & 0.587 \\
	 4 		           & 0.646 & 0.733 & 0.804 & 0.724 & 0.615 & 0.581 \\
	 8 		           & 0.649 & 0.745 & 0.810 & 0.707 & 0.626 & 0.636  \\
	 16 		       & 0.649 & 0.748 & 0.814 & 0.732 & 0.638 & 0.615  \\
    \bottomrule
    \end{tabular}
\end{table}

\begin{figure}
	\centering
	\includegraphics[width=\linewidth]{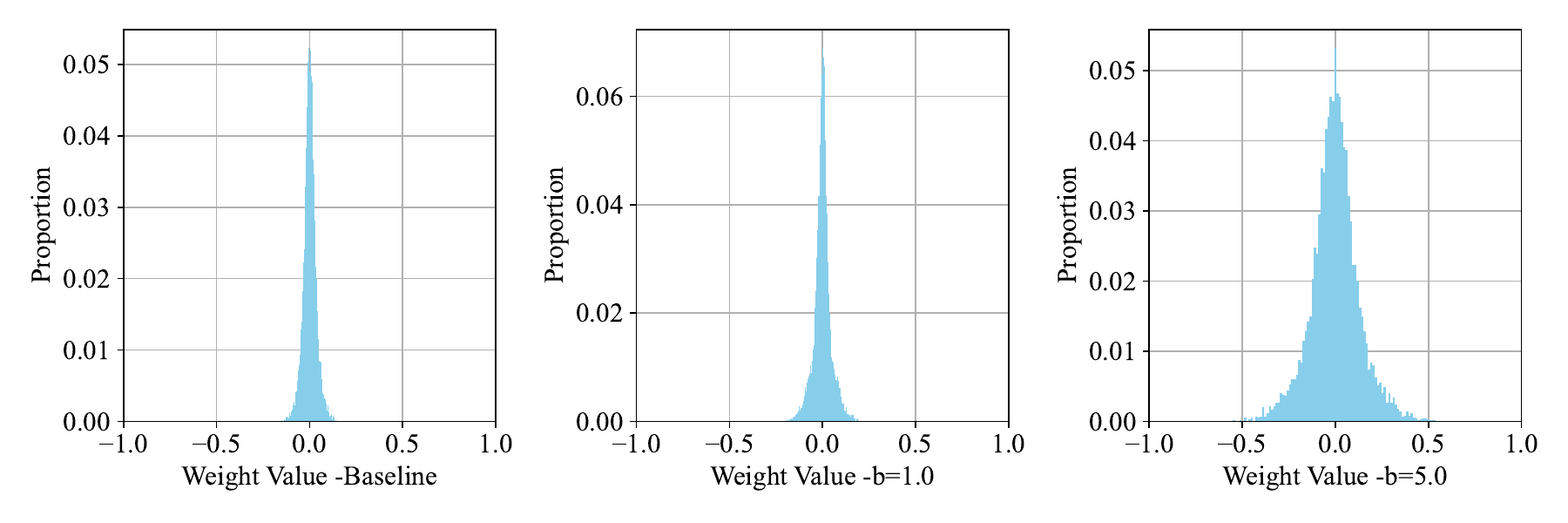}
	\caption{Parameter distribution of watermarked StyleGAN2 for two different values of $b$. Baseline stands for the original (non watermarked) model.}
	\label{fig:weigt distribution}
\end{figure}

We also evaluated the influence of varying $b$ and $k$ on the robustness of the watermark against fine-tuning attacks (see Table~\ref{tab: ft k and b}).
We can observe that when $k$ is greater than 4, further increments in $k$ do not significantly improve the robustness. However, as $b$ increases, the robustness initially strengthens, followed by a sharp decline, and the optimal robustness is reached roughly when $b$ equals 1.0.
We hypothesize that this is due to the fact that i) an overly large $b$ results in lower Bit Acc before the attack, and ii) an overly large $b$ causes a larger change in the distribution of the model weights compared to the original non-watermarked model (see Fig.~\ref{fig:weigt distribution}). This leads to larger gradients ( $ \partial L_G / \partial \theta$), causing more drastic changes of weights during the fine-tuning attack and a faster escape from the wide flat minimum.

The above results can provide valuable guidance for choosing appropriate $k$ and $b$. On one hand, a rather small $k$ is enough to achieve a good balance between fidelity, Bit Acc, and robustness, while also reducing computational cost of the embedding procedure. On the other hand, we believe that the key to achieving improved robustness is to choose a suitable $b$ for the wide flat minimum search algorithm, making it sufficiently wide and inducing minimal changes to the distribution of model weights.

\section{Conclusion}
\label{sec: CONCLUSION}

With the increasing commercial value of image generation models, IPR protection has become an essential problem. In this paper, we have proposed a robust box-free watermarking method, based on WFM search, to protect the IPR of GANs and image processing networks. Extensive experiments demonstrate that our method achieves significant robustness in different white-box model-level attacks and image processing attacks.

Our work indicates that the robustness of existing box-free GAN watermarking methods has flaws, as the watermark can be easily removed through fine-tuning attacks. We attribute this to the unique nature of box-free methods, which essentially work by fitting the distribution of watermarked data. Thus using clean data for fine-tuning attacks can shift the generator's distribution, resulting in the removal of the watermark. Essentially, the key to improving the robustness of model-level attacks lies in improving the robustness of watermarking loss to parameter perturbations of the generator. In our method, we do so by treating watermark embedding and generator training as two distinct tasks, and looking for a wide flat minimum of the embedding loss to prevent the model from forgetting the watermark during fine-tuning attacks.
Our experiments also indicate that the key to achieving optimal robustness is that the generator converges to a minimum as wide as possible while the parameter distribution of the watermarked generator is close to that of the non-watermarked generator.

An interesting direction for future research is to extend our method to more advanced generative models, such as diffusion models (DM). DMs are gradually replacing GANs in an increasing amount of practical applications. Hence, protecting the IPR of DMs is becoming an important issue. The training and image generation process of DMs are different from GANs, thus calling for specific designs.

\bibliographystyle{IEEEtran}
\bibliography{adaptivev2}

\end{document}